\documentclass[runningheads]{llncs}
\usepackage[T1]{fontenc}
\usepackage{graphicx}
\usepackage{booktabs}
\usepackage[misc]{ifsym}

\usepackage{mwe}
\usepackage{amsmath}
\usepackage[ruled,vlined, linesnumbered]{algorithm2e}
\usepackage{amssymb}
\usepackage{graphicx}
\usepackage{subcaption}
\usepackage{multirow} 
\usepackage{hyperref}
\usepackage{xcolor}
\makeatletter

\begin{document}

\title{GEAR: Geography-knowledge Enhanced Analog Recognition Framework in Extreme Environments}

\author{
Zelin Liu\textsuperscript{*}\enspace,
Bocheng Li\textsuperscript{*}\enspace,
Yuling Zhou\textsuperscript{*}\enspace,
Xuanting Li, \\
Yixuan Yang, \textbf{Jing Wang\textsuperscript{\dag},
Weishu Zhao\textsuperscript{\dag}, Xiaofeng Gao\textsuperscript{\dag}}\\
Shanghai Jiao Tong University
}
\begingroup
\renewcommand\thefootnote{\fnsymbol{footnote}} 
\footnotetext[1]{Equal contribution.}
\footnotetext[4]{Corresponding authors: zhenjiaofenjie@sjtu.edu.cn, zwsh88@sjtu.edu.cn, gao-xf@cs.sjtu.edu.cn.}
\endgroup

\institute{}

\maketitle              

\begin{abstract}
The Mariana Trench and the Qinghai-Tibet Plateau exhibit significant similarities in geological origins and microbial metabolic functions. Given that deep-sea biological sampling faces prohibitive costs, recognizing structurally homologous terrestrial analogs of the Mariana Trench on the Qinghai-Tibet Plateau is of great significance. 
Yet, no existing model adequately addresses cross-domain topographic similarity retrieval, either neglecting geographical knowledge or sacrificing computational efficiency.
To address these challenges, we present \underline{\textbf{G}}eography-knowledge \underline{\textbf{E}}nhanced \underline{\textbf{A}}nalog \underline{\textbf{R}}ecognition (\textbf{GEAR}) Framework, a  three-stage pipeline designed to efficiently retrieve analogs from 2.5 million square kilometers of the Qinghai-Tibet Plateau: 
(1) Skeleton guided Screening and Clipping: Recognition of candidate valleys and initial screening based on size and linear morphological criteria.
(2) Physics aware Filtering: The Topographic Waveform Comparator (TWC) and Morphological Texture Module (MTM) evaluate the waveform and texture and filter out inconsistent candidate valleys.
(3) Graph based Fine Recognition: We design a \underline{\textbf{M}}orphology-integrated \underline{\textbf{S}}iamese \underline{\textbf{G}}raph \underline{\textbf{N}}etwork (\textbf{MSG-Net}) based on geomorphological metrics.
Correspondingly, we release an expert-annotated topographic similarity dataset targeting tectonic collision zones.
Experiments demonstrate the effectiveness of every stage. Besides, MSG-Net achieved an F1-Score 1.38 percentage points higher than the SOTA baseline. 
Using features extracted by MSG-Net, we discovered a significant correlation with biological data, providing evidence for future biological analysis.

\end{abstract}

\section{Introduction}

\begin{figure*}[!htb]
    \centering
    \includegraphics[width=0.9\linewidth]{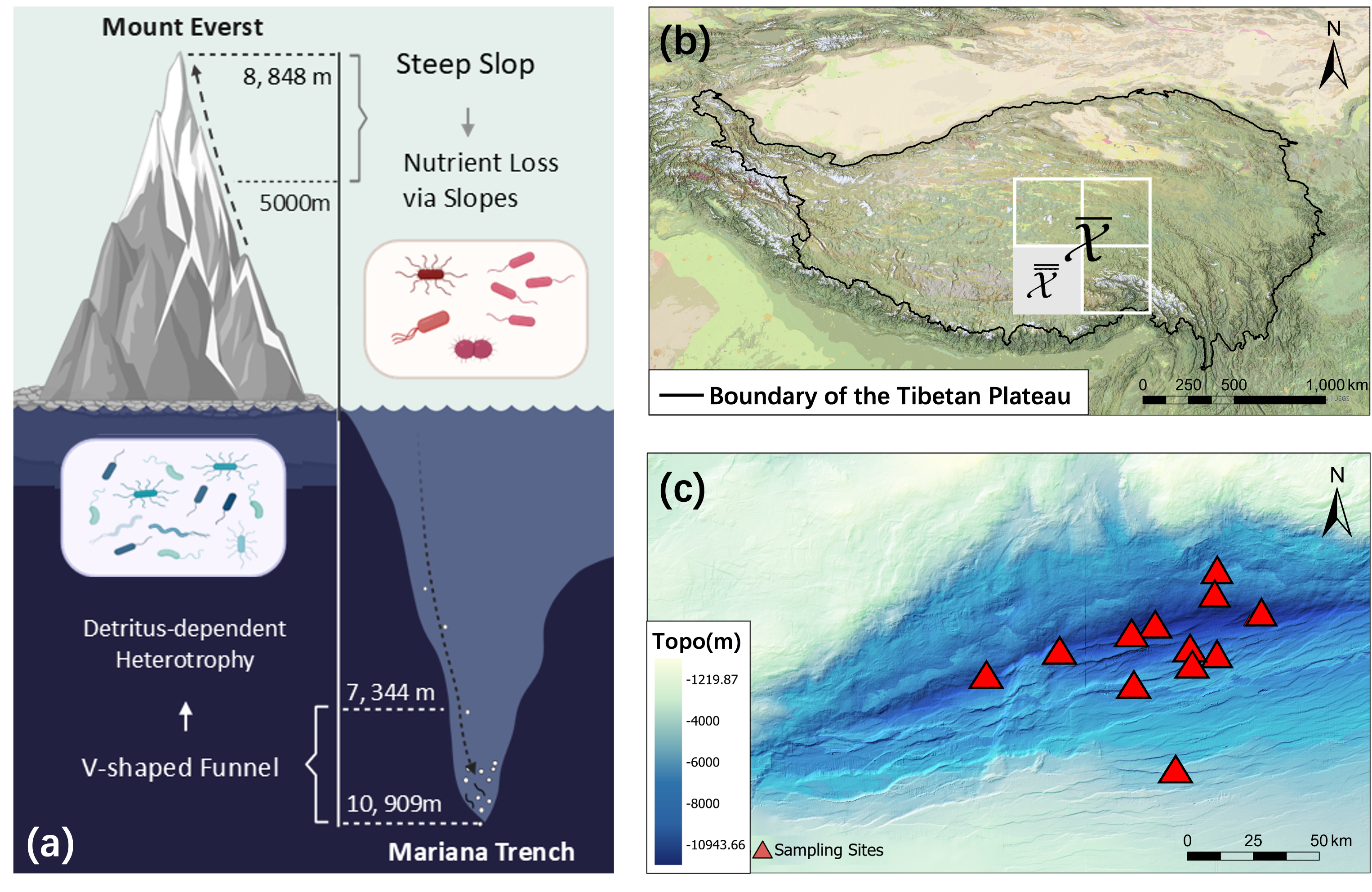}
    \caption{ Topology and microbial comparison between the Tibetan Plateau (TP) and the Mariana Trench (MT).(a) V-shaped topographic profile comparison.(b) Location and DEM grids of the TP.(c) Location of the MT and sampling sites.}
    \label{fig.1}
\end{figure*}

The Qinghai-Tibet Plateau and the Mariana Trench share structural origins driven by tectonic collision and exhibit over 90\% similarity in functional metabolic modules of their microbial communities\cite{stern2002subduction,jamieson2010hadal,liu2022comparison}. This similarity may be caused by their shared V-shaped topography (Fig.~\ref{fig.1}(a)), in which local slopes influence nutrient accumulation through gravity-driven processes. As deep-sea biological sampling faces prohibitive costs, identifying structurally homologous terrestrial analogs on the Qinghai-Tibet Plateau provides a promising way to roughly estimate the biological conditions of certain regions in the Mariana Trench and guide the selection of future sampling sites. Thus, a framework to recognize terrestrial analogs across the Qinghai-Tibet Plateau is required.

However, existing methods present limitations in measuring cross-domain topographic similarities.
Firstly, traditional geographic methods lack strong representational capacity~\cite{pike2009geomorphometry}.
Secondly, Computer Vision (CV)-based approaches often disregard 3D structural trends when processing Digital Elevation Models (DEMs)~\cite{zhu2017deep}.
Thirdly, recent Geospatial AI (GeoAI) models, such as GCN-DP~\cite{kong2025integrating}, remain primarily designed for specific regional analysis and lack the ability to support terrain matching across domains.

To address these limitations, we propose the \underline{\textbf{G}}eography-knowledge \underline{\textbf{E}}nhanced \underline{\textbf{A}}nalog \underline{\textbf{R}}ecognition (\textbf{GEAR}) Framework, a three-stage pipeline designed to recognize valleys with similar sizes and structures to a reference trench in the Mariana Trench from an area of 2.5 million square kilometers of the Qinghai-Tibet Plateau (QTP):

(1) \textbf{Coarse Screening}: We extract approximately 30,000 candidate valleys in the QTP via a proposed Skeleton-guided Screening and Clipping (SSC) approach. Specifically, we proposed a Local Elevation Deficit Binarization (LEDB) strategy to identify valley areas. Using the Zhang-Suen thinning algorithm \cite{zhang1984fast}, we extract the skeletons of these areas. Since the Mariana Trench exhibits a straight morphology under regional tectonic stress fields\cite{zoback1992first,molnar1977collision}, we apply linear regression fitting to extract landforms with similar straightness and crop the corresponding rectangular regions.

(2) \textbf{Physics-aware Filtering:} To match overall topographies efficiently, we introduce the Topographic Waveform Comparator (TWC) and the Morphological Texture Module (MTM). We first decompose the candidate valleys into slice sequences. TWC measure structural similarity via Derivative Dynamic Time Warping (DDTW\cite{keogh2001derivative}) based on morphological trends. Subsequently, MTM extracts terrain texture by Eigenshape Analysis\cite{macleod1999generalizing}, ranking candidates by cosine similarity. This stage efficiently compresses the candidate pool from roughly $10^4$ to $10^3$.

(3) \textbf{Graph-based Fine Recognition}: A Morphology-integrated Siamese Graph Network (MSG-Net) is employed to calculate the topological similarity between candidate valleys and the reference trench through shared-weight graph convolutions incorporating traditional topographic knowledge. This process resolves the topological distance between samples, retrieving most similar candidate valley to the reference Trench.

This architecture resolves the limitations of existing methods. First, to overcome the limited capacity of traditional indices, GEAR captures the overall morphological structure. Second, unlike CV approaches that process DEMs as planar grids, the pipeline preserves physical topography by integrating geographical priors. Third, diverging from GeoAI classification, our matching process measures structural similarities across extremely different heights.

Our key contributions are summarized as follows:
(1) We achieve fast, topography-aware cross-regional terrain analog retrieval in extreme environments, overcoming the 2D flattening of CV models and the extreme height barriers that fail current GeoAI methods.
(2) We introduce an annotated open-source topographic similarity dataset targeting tectonic collision zones\footnote{The dataset and source code are provided for double-blind review at: \url{https://anonymous.4open.science/r/GEAR-Blind}}. Experimental evaluations indicate that GEAR achieves an accuracy of 83.93\%, a recall of 98.05\%, and an F1-score of 86.04\%, outperforming the highest performing baseline, SANI-SSL (which achieved 79.82\% accuracy and 86.49\% recall).
(3) We validate the topography-microbiome correlation in the Mariana Trench and Mount Everest, establishing an interdisciplinary methodology that maps macro-geomorphology to biological constraints. Utilizing AI-matched QTP analogs, we provide predictive, actionable guidance for deep-sea ecological sampling and functional transfer research.\cite{jamieson2010hadal,liu2022comparison}.


\section{Problem Formulation}

\subsection{Geographical Analysis and Data Processing}
Geographical analysis focuses on two extreme environmental regions: (1) the Qinghai-Tibet Plateau (QTP), and (2) the Mariana Trench.

\paragraph{\textbf{Qinghai-Tibet Plateau (QTP).}} The QTP region is between latitude 67$^\circ$40 E $\sim$ 104$^\circ$40 E and longitude 25$^\circ$59 N $\sim$ 40$^\circ$01 N with an irregular boundary. We use the Digital Elevation Model (DEM) data from the ASTER GDEM dataset\footnote{\url{https://data.tpdc.ac.cn/zh-hans/data/bdacc6b7-aff0-49b1-b666-cb97d846d072}}, which comprises 922 coordinate-based grids in .tif format, denoted as $\mathcal{\overline{\overline{X}}} = \{\overline{\overline{X}}_1, \overline{\overline{X}}_2, \dots, \overline{\overline{X}}_{m''}\}$, and $m''=922$. Each \emph{grid} $\overline{\overline{X}}_i$, depicts a unit sub-region of latitude $1^\circ$ $\times$ longitude $1^\circ$, has a size of 24 MB, with a spatial resolution of 30$m$. $\overline{\overline{X}}_i$ consists of 3601 $\times$ 3601 pixels, whose value represents the elevation of the corresponding sub-region of $30m \times 30m$, as shown in Fig.~\ref{fig.1}(B). 

To avoid mismatches along grid boundaries, we concatenate $\mathcal{\overline{\overline{X}}}$ into a new dataset $\mathcal{\overline{X}} = \{\overline{X}_1, \overline{X}_2, \dots, \overline{X}_{m'}\}$. Each $\overline{X}_j$, denoted as a \emph{tile}, is composed of $2 \times 2$ neighboring $\overline{\overline{X}}_i$'s with zero-padding borders. Adjacent tiles $\overline{X}_{j}$ and $\overline{X}_{j+1}$ share two duplicated grids; thus, $m'=m''=922$. We further extract valleys from the DEM tiles $\mathcal{\overline{X}}$, denoted as $\overline{\overline{\mathcal{V}}}=\{\overline{\overline{V}}_1, \cdots, \overline{\overline{V}}_{m}\}$. Each valley $\overline{\overline{V}}_i$ has a region of $5 km \sim 40  km$, representing a topographic valley in real world. %

\paragraph{\textbf{Mariana Trench (MT).}} The DEM data for the Mariana Trench is a strict square region between  140.5963$^\circ$E$\sim$143.1206$^\circ$E and 10.5224$^\circ$N$\sim$11.808$^\circ$N, extracted from the GEBCO 2024 dataset\footnote{\url{https://www.gebco.net/data-products/gridded-bathymetry-data}}, with a spatial resolution of 100$m$ (and a size of 372 KB). As shown in Fig.~\ref{fig.1}(C), we also use 790 deep-sea microbial samples from 46 valid in-situ sampling points gathered by China's ``Fendouzhe'' manned submersible at the bottom of the Mariana Trench from February to October 2021. Fig.~\ref{fig.1}(C) depicts the position of sampling points as red triangles.

\begin{figure}[htbp]
    \centering
    \includegraphics[width=\textwidth]{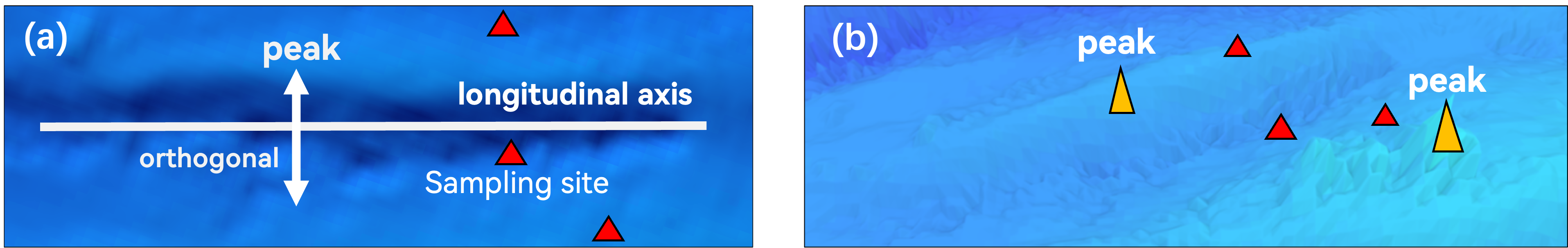} 
    \caption{The reference trench $\mathcal{T}_{\text{ref}}$ (2D (a) and 3D (b) versions)}
    \label{fig:truth}
\end{figure}

To achieve analog recognition between QTP and MT, we select a \emph{reference trench} $\mathcal{T}_{\text{ref}}$ from MT via two steps: (1) Centered from a sampling point, which is the lowest position of a trench, we fix the longitudinal axis of $\mathcal{T}_{\text{ref}}$ according to the geological mechanisms of oceanic plate subduction and compression~\cite{Zhang230235}; and (2) we bound $\mathcal{T}_{\text{ref}}$ using the orthogonal direction to the longitudinal axis, climbing upward along the slopes on both sides of the trench to find the peak. Finally, $\mathcal{T}_{\text{ref}}$ is a $25km \times 3km$ rectangular area, as shown in Fig.~\ref{fig:truth}.







\paragraph{\textbf{Expert-Annotated Dataset.}} We adopt supervised learning with an expert-annotated dataset to quantify topographic Similarity between pairwise valleys and trenches. The labels in this dataset are binary: 1 denotes similarity, while 0 denotes dissimilarity. It comprises 305 high-quality sample pairs, strictly balanced between 154 positive pairs and 151 negative pairs. We conduct the Cohen's kappa test~\cite{landis1977measurement} to assess inter-rater reliability, yielding a $K$-value of 0.698, indicating substantial agreement among the experts.



\subsection{Problem Definition}



\begin{definition}[Similarity Learning Model]
Given a QTP candidate valley $V_i$ and a trench $T_j$, a learning model $f$ is trained by the Expert-Annotated Dataset and predicts the similarity score, say, $f(V_i, T_j) \in (0,1]$.
\end{definition}

We employ the Binary Cross-Entropy (BCE) loss function for the model's loss function. A similarity score closer to 1 indicates a higher degree of similarity in spatial topographical structure between the two terrains.

\begin{definition}[Analog Recognition Retriever]
Given a QTP candidate valley set $\overline{\overline{\mathcal{V}}}=\{\overline{\overline{V}}_1, \cdots, \overline{\overline{V}}_{m}\}$, and a reference trench $\mathcal{T}_{\text{ref}}$, an Analog Recognition Retriever selects the most similar valley w.r.t. $\mathcal{T}_{\text{ref}}$, say, $\arg\max_i f(\overline{\overline{V}}_i, \mathcal{T}_{\text{ref}})$. 

\end{definition}



\section{GEAR Framework}
A coarse-to-fine topographic retrieval framework, GEAR, is developed to match the reference trench in Mariana with analogous valleys across the Qinghai-Tibet Plateau (QTP). Initial valley candidates are first extracted using a Skeleton-guided Screening and Clipping (SSC) approach based on tectonic linearity. The Topographic Waveform Comparator (TWC) and Morphological Texture Module (MTM) are subsequently applied to evaluate cross-sectional profiles and filter the candidates to the top 1,000 matches. Finally, a Morphology-integrated Siamese Graph Network (MSG-Net) computes the deep topological similarities to identify the most homologous valleys.

\begin{figure}[htbp]
    \centering
    \includegraphics[width=\textwidth]{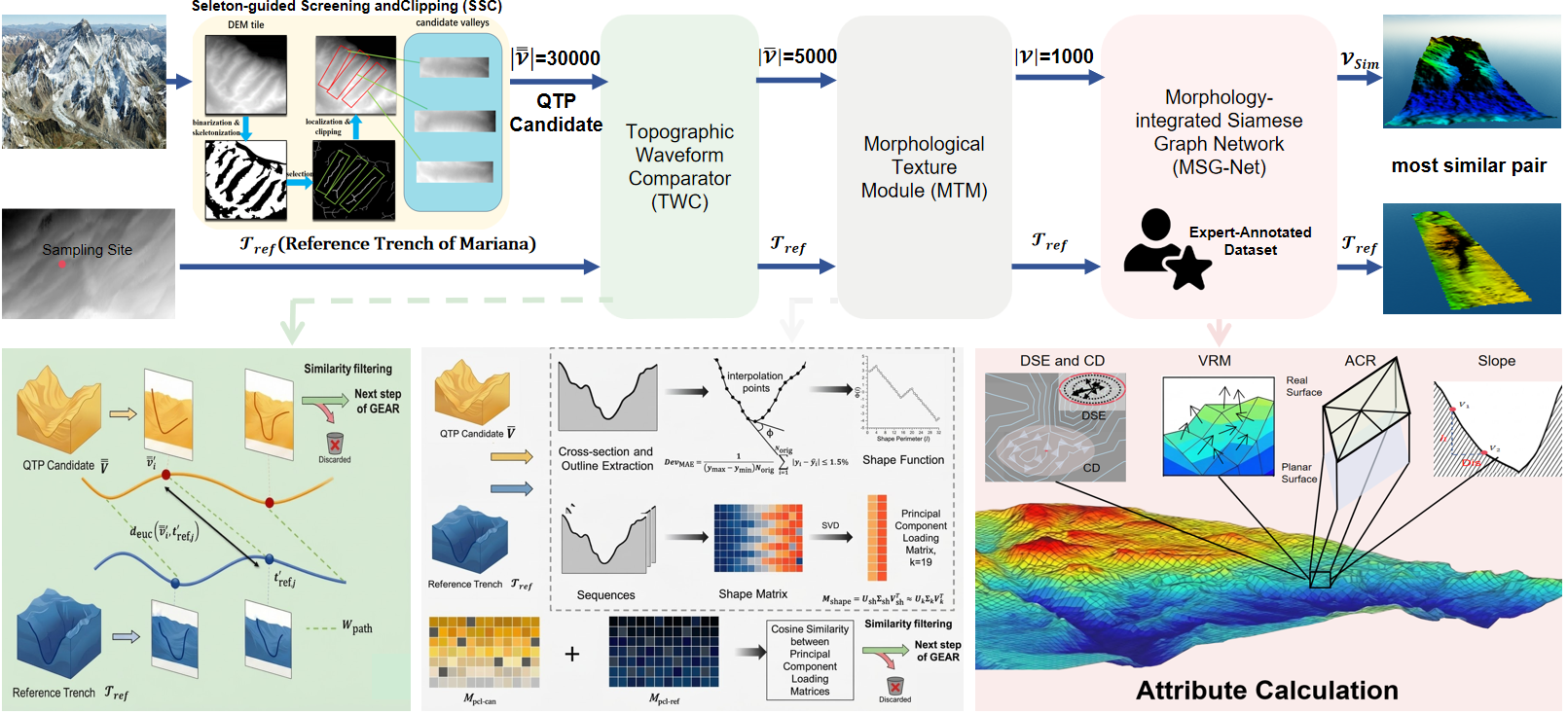} 
    \caption{The pipeline of GEAR}
    \label{fig:pipeline}
\end{figure}

\subsection{Seleton-guided Screening and
Clipping (SSC) }
\indent As the first-stage coarse screening step, we propose a Skeleton-guided Screening and Clipping (SSC) algorithm considering the prior knowledge of the straight morphology of the Mariana Trench. The detailed workflow consists of four steps, as shown in Alg.~\ref{algo:trench_extraction}.






\begin{algorithm}[htbp]
\caption{Skeleton-guided Screening and Clipping (SSC)}
\label{algo:trench_extraction}
\SetKwInput{KwIn}{Input}
\SetKwInput{KwOut}{Output}
\SetKwFunction{LS}{LS}  
\SetKwFunction{elev}{elev}  

\KwIn{DEM tiles $\mathcal{\overline{X}} = \{\overline{X}_1, \dots, \overline{X}_{m'}\}$, 
    neighborhood size $blk$, 
    elevation threshold $c$, 
    line-error threshold $err$, segment length thresholds $s_l$, $s_u$. 
}
\KwOut{
    Set of candidate valleys  $\mathcal{\overline{\overline{V}}} = \{\overline{\overline{V}}_1, \dots, \overline{\overline{V}}_{m}\}$ 
}

$\mathcal{\overline{\overline{V}}} \leftarrow \emptyset$; $k \leftarrow 1$\; 
\For{$t = 1$ \textbf{to} $m'$} {  
\tcp{\textcolor{blue}{1. Local Elevation Deficit Binarization (LEDB)}}
    \ForEach{pixel $p \in \overline{X}_t$}{
            $\mu_p$ = mean elevation for $blk$-size neighborhood of $p$\;  
            $I(p) = \begin{cases}1, & \text{elevation value of } p < \mu_p - c \\ 0, & \text{otherwise}\end{cases}$ \;  
        }
        $P_{\text{valley}} \leftarrow \{p \in \overline{X}_t \mid I(p) = 1\}$ \;  

    \tcp{\textcolor{blue}{2. Skeleton Extraction}}

     Extract skeletons via Zhang-Suen thinning algorithm on $P_{\text{valley}}$\;
     
     Seperate connected components of skeletons as $\mathcal{S}_t = \{S_{t,1}, \dots, S_{t,k_t}\}$\; 

    \tcp{\textcolor{blue}{3. Linear Feature Selection}}
        \ForEach{$S_{t,i} \in \mathcal{S}_t$} {
            Fit a segment of $S_{t,i}$ via LR with slope $a$, intercept $b$, length $len$\; 
            Compute MSE error $mse$ between the segment and $S_{t,i}$\;
            \tcp{\textcolor{blue}{4. Target Region Localization and Clipping}}
            \If{$mse \le err$ \text{and} $s_l \le len \le s_u$}{
                Compute bounding box of $S_{t,i}$, crop sub-region $\overline{\overline{V}}_k$ from $\overline{X}_t$\;
                $\mathcal{\overline{\overline{V}}} \leftarrow \mathcal{\overline{\overline{V}}} \cup \{\overline{\overline{V}}_k\}$; $k \leftarrow k + 1$\;
            }
        }
}
\Return{$\mathcal{\overline{\overline{V}}}$} \; 
\end{algorithm}

After removing duplicate valleys, approximately 30,000 candidate valleys were selected in the QTP via the SSC algorithm.

\subsection{Topographic Waveform Comparator (TWC)}
We design the Topographic Waveform Comparator (TWC) as a coarse filter to capture the spatial waveform from the candidate valley set $\overline{\overline{\mathcal{V}}}$. First, we decompose candidate valleys $\overline{\overline{V}} \in \overline{\overline{\mathcal{V}}}$ and the reference trench $\mathcal{T}_{\text{ref}}$ into multivariate slice sequences. To unify vector dimensions for point-wise Euclidean comparisons, $\mathcal{T}_{\text{ref}}$ is down-sampled to align with the candidates' uniform 38-point width. This transforms the terrain blocks into varying-length slice sequences, denoted as $\overline{\overline{v}} = (\overline{\overline{v}}_1, \dots, \overline{\overline{v}}_n)$ and $t_{\text{ref}} = (t_{\text{ref},1}, \dots, t_{\text{ref},m})$.

Because standard DTW aligns absolute coordinates and fails under extreme vertical elevation shifts, we apply Derivative Dynamic Time Warping (DDTW) \cite{keogh2001derivative} to measure similarities based on morphological trends. The structural proximity of the sequence derivatives is evaluated in $\mathcal{O}(L^2)$ time, where $L$ is the sequence length:
\begin{equation}
Cost_{\text{DDTW}}(\overline{\overline{v}}, t_{\text{ref}}) = \min_{W_{\text{path}}} \sum_{(i,j) \in W_{\text{path}}} d_{\text{euc}}(\overline{\overline{v}}'_i, t'_{\text{ref},j})
\end{equation}
where $W_{\text{path}}$ is the warping path, and $d_{\text{euc}}$ denotes the Euclidean distance between first-order slice derivatives. To further ensure orientation invariance, we perform bidirectional DDTW comparisons. Notably, unconstrained DDTW is explicitly chosen over penalized variants (e.g., Amerced-DTW \cite{Itakura1162641}) to prioritize retrieval recall, purposefully allowing matches with large spatial stretching. This stage forms the filtered top-5,000 candidate set $\overline{\mathcal{V}} = \{\overline{V}_1, \dots, \overline{V}_{k'}\}$.

\subsection{Morphological Texture Module (MTM)}
We design the Morphological Texture Module (MTM) to evaluate the topographic texture for the TWC-screened candidate set $\overline{\mathcal{V}}$. This module executes an extended Eigenshape Analysis \cite{lohmann1983eigenshape} \cite{macleod1999generalizing} on the slice sequences obtained in TWC.

To extract morphological features via Eigenshape analysis, slice sequences must be uniformized. Rather than utilizing strict length-preserving interpolation, we introduce a Mean Absolute Error (MAE) constrained down-sampling. This prevents overfitting to micro-topographic noise while successfully preserving macro-scale geometries. For each original slice sequence, the threshold is defined as:
\begin{equation}
Dev_{\text{MAE}} = \frac{1}{(y_{\text{max}} - y_{\text{min}}) N_{\text{orig}}} \sum_{i=1}^{N_{\text{orig}}} \left| y_{i} - \hat{y}_{i} \right| \leq 1.5\%
\end{equation}
where $y_{i}$ and $\hat{y}_{i}$ represent original and reconstructed depths, and $y_{\text{max}} - y_{\text{min}}$ denotes the elevation span. By adhering to this bounded deviation, the reference trench $\mathcal{T}_{\text{ref}}$ slices (originally 216 points) are down-sampled into 181 equidistant coordinates. Consequently, all candidate slices $\overline{v}$ of $\overline{V} \in \overline{\mathcal{V}}$ are up-sampled to match this 181-coordinate baseline.

Following this uniform discretization, shape functions are computed by measuring the angles of the line segments connecting adjacent interpolated points. This angular representation assembles a characteristic $m \times n$ shape matrix $M_{\text{shape}}$. In $\mathcal{O}(\min(m^2n, mn^2))$ time, spatial features are extracted via Singular Value Decomposition (SVD):
\begin{equation}
M_{\text{shape}} = U_{\text{sh}} \Sigma_{\text{sh}} V_{\text{sh}}^T \approx U_{k} \Sigma_{k} V_{k}^T
\end{equation}
Based on the $\mathcal{T}_{\text{ref}}$ decomposition, we truncate the system to the top $k_{\text{pc}}=19$ principal components, capturing $>80\%$ of its morphological variance. Using this exact $k_{\text{pc}}$ configuration, we generate the $19 \times 179$ principal component loading matrices $M_{\text{pcl}}$ for all candidates. Finally, candidates are ranked against $\mathcal{T}_{\text{ref}}$ via the cosine similarity of their respective loading matrices, refining the intermediate pool $\overline{\mathcal{V}}$ to the final 1,000 top-tier samples $\mathcal{V} = \{V_1, \dots, V_k\}$.

\subsection{Morphology-integrated Siamese Graph Network (MSG-Net)}
To accurately quantify the topological homology between cross-domain extreme environments, we propose the Morphology-integrated Siamese Graph Network (MSG-Net). 
Prior to deep feature extraction, candidate valleys ($\{V_1, V_2, \dots, V_k\}$) and reference trench ($\mathcal{T}_{\text{ref}}$) received from MTM are transformed into graph structures $G(N, E, A)$, where $N$ denotes the set of nodes, $E$ denotes the set of edges based on spatial proximity, and $A$ is the adjacency matrix. Nodes are generated by equidistantly sampling the original contour lines, then the Delaunay triangulation algorithm is applied to these nodes to establish the edges, thereby constructing a spatial network that forms the topological skeleton $E$.

Focusing on each node in a valley or trench, we adopt five geographic metrics to evaluate their similarity. 

\textbf{Vector Ruggedness Measure (VRM)~\cite{sappington2007}}: Quantifies terrain fragmentation by calculating the dispersion degree of 3D normal vectors across adjacent pixels. For each pixel $p_i$, define its normal vector as $\overrightarrow{p_i}$, the VRM for a specific neighborhood with $n$ pixels (e.g., $3 \times 3$ window) is defined as:
    \begin{equation}
    \text{VRM} = 1 - \frac{1}{n}\sqrt{\left|\sum\nolimits_{i=1}^{n} \overrightarrow{p_i}\right|^2 }.
    \end{equation}
    
\textbf{Arc-Chord Ratio (ACR)~\cite{dupreez2015}}: decouples surface roughness from macroscopic slope by calculating the ratio of the true contoured surface area to the planar projection area on the plane of best fit, estimating the absolute physical space available for benthic attachment. It is formulated as:
    \begin{equation}
        \text{ACR} = \frac{Area_{\text{surface}}}{Area_{\text{planar}}}.
    \end{equation}
where $Area_{\text{surface}}$ is the actual 3D contoured surface area, and $Area_{\text{planar}}$ is the 2D projected area of the terrain patch on the optimal plane of best fit.
    
\textbf{Slope}: Derived by calculating the arctangent of the elevation difference divided by the horizontal distance, representing the macroscopic gravitational and fluid-dynamic gradients. For adjacent nodes $i$ and $j$, it is computed as:
    \begin{equation}
        \text{Slope}_{i,j} = \arctan\left(\frac{|\Delta h_{i,j}|}{Dis_{i,j}}\right).
    \end{equation}
where $\Delta h_{i,j}$ is the absolute elevation difference and $Dis_{i,j}$ is the 2D horizontal Euclidean distance between the nodes.
    
\textbf{Contour Density (CD)~\cite{kong2025integrating}}: This metric is defined as the concentration of contour lines within a predefined local neighborhood of radius $r$(we set it as 500 meter) around each node, formulated as:
\begin{equation}
    CD = \frac{\sum_{i=1}^{m} L_i}{\pi r^2}
\end{equation}
where $L_i$ denotes the length of the $i$-th contour segment within the circular search area, and $m$ is the total number of segments.

\textbf{Direction Shannon Entropy (DSE)~\cite{kong2025integrating}}: The azimuths of contour segments within the radius $r$ are discretized into $N$(we set it as 36) uniform angular bins. The entropy is calculated as:
\begin{equation}
    DSE = -\sum_{j=1}^{N} p_j \ln(p_j)
\end{equation}
where $p_j$ represents the probability of contour segments falling into the $j$-th directional bin. 

To maintain numerical stability during information propagation, the normalized graph Laplacian matrix $L = I_m - D^{-1/2} A D^{-1/2}$ is computed to replace the raw adjacency matrix $A$, ensuring stable feature aggregation.
MSG-Net formulates the cross-environment comparison as a metric-learning task using a parameter-sharing Siamese architecture. The constructed graphs are first processed by a hierarchical Graph Convolutional Network (GCN) backbone. For the input sample pair, the reference trench graph ($\text{Graph}_1$) and the candidate valley graph ($\text{Graph}_2$) are fed into the identical GCN backbone, projecting them into a unified high-dimensional latent space to yield topological representations $emb_1$ and $emb_2$, respectively. Subsequently, the model calculates the absolute divergence feature between the two terrain:

\begin{equation}
    F_{diff} = |emb_1 - emb_2|
\end{equation}

Finally, $F_{diff}$ is input into a bottleneck Multi-Layer Perceptron (MLP) discriminator. Utilizing BatchNorm, ReLU activation, and Dropout mechanism to prevent overfitting, the discriminator outputs logits that are passed through a Sigmoid function, ultimately mapping to a terrain ecological similarity probability ranging from 0 to 1.

The model is initially trained on the expert-annotated dataset described in Section 2.1. Once optimized, the parameters are frozen, then it is employed to compute the similarity score. The candidate valley exhibiting the highest similarity to reference trench is identified as the final output of our pipeline.

\section{Experiments}
\subsection{Implementation Details}
Models were trained on NVIDIA A100 GPUs under CUDA 13.0. The architecture comprises a 3-layer GCN (128D hidden/output units, 0.1 pooling ratio) and a 3-layer MLP classifier (256 and 64 hidden units). Dropout rates for the GCN and MLP were set to 0.3 and 0.5, respectively. We optimized the BCE loss using Adam (learning rate: $10^{-4}$, weight decay: $10^{-3}$). Training was executed with a batch size of 32, gradient clipping at 2.0, and early stopping.

\subsection{Baseline Setting}
To evaluate MSG-Net, we benchmark against eight representative methods categorized into three groups (Table \ref{tab:main_results}):

\textbf{Traditional ML}:  Includes classical classifiers relying on raw elevation or handcrafted features: Random Forest (RF) \cite{belgiu2016}, Support Vector Machine (SVM) \cite{mountrakis2011}, and PCA-EPFs \cite{kang2017}, a spatial-spectral baseline combining PCA with edge-preserving filters.
    
\textbf{Deep Learning}: We include a standard 2D-CNN \cite{hu2015}, a hybrid MLP-CNN architecture fusing 1D attributes and 2D spatial features, and the ResNet50 \cite{he2016}. Furthermore, SANI-SSL \cite{wang2020} is introduced as a spatial-aware paradigm that exploits shape-adaptive neighborhood information.
    
\textbf{Geo-aware Model}: Alongside our MSG-Net, we benchmark against GCN-DP \cite{kong2025integrating}, a robust Graph Convolutional Network integrated with differentiable pooling. While it similarly relies on graph representations, its node attributes are strictly limited to contour-specific metrics, lacking integration of traditional topographic knowledge. It serves as a direct topological baseline to validate the superiority of our MSG-Net.

\subsection{Experiment Results}
\textbf{Comparative Evaluation of Model Performance:}
\begin{table*}[h]
\centering
\caption{Comparative experiment result}
\label{tab:main_results}
\resizebox{\textwidth}{!}{
\begin{tabular}{llcccc}
\toprule
\textbf{Category} & \textbf{Method} & \textbf{Accuracy (\%)} & \textbf{Precision (\%)} & \textbf{Recall (\%)} & \textbf{F1-Score (\%)} \\
\midrule
\multirow{3}{*}{Traditional ML} 
& RF & 64.87 $\pm$ 0.94 & 66.58 $\pm$ 1.51 & 64.87 $\pm$ 0.94 & 61.93 $\pm$ 1.21 \\
& SVM & 63.18 $\pm$ 0.65 & 56.93 $\pm$ 1.40 & 63.18 $\pm$ 0.66 & 56.67 $\pm$ 0.83 \\
& PCA-EPFs & 68.45 $\pm$ 1.02 & 69.12 $\pm$ 1.33 & 66.83 $\pm$ 1.15 & 67.95 $\pm$ 1.24 \\
\midrule

\multirow{4}{*}{Deep Learning} 
& 2D-CNN & 73.52 $\pm$ 1.15 & 70.15 $\pm$ 1.48 & 78.31 $\pm$ 1.22 & 73.98 $\pm$ 1.30 \\
& MLP-CNNs & 74.81 $\pm$ 1.20 & 71.33 $\pm$ 1.52 & 80.14 $\pm$ 1.18 & 75.47 $\pm$ 1.35 \\
& ResNet50 & 77.63 $\pm$ 1.11 & 75.88 $\pm$ 1.41 & 82.05 $\pm$ 1.09 & 78.84 $\pm$ 1.25 \\
& SANI-SSL & 79.82 $\pm$ 1.08 & 77.54 $\pm$ 1.38 & 86.49 $\pm$ 0.96 & 81.79 $\pm$ 1.17 \\
\midrule

Geo-aware Model
& GCN-DP & 81.57 $\pm$ 1.24 & \textbf{79.71 $\pm$ 1.09} & 90.28 $\pm$ 1.37 & 84.66 $\pm$ 1.58 \\

& \textbf{MSG-Net (Ours)} & \textbf{83.93 $\pm$ 1.05} & 76.65 $\pm$ 1.15 & \textbf{98.05 $\pm$ 0.85} & \textbf{86.04 $\pm$ 1.13} \\
\bottomrule
\end{tabular}%
}
\end{table*}
During the experiment, we used a 5-fold cross-validation strategy to ensure robust generalization, and the model achieved an F1-score of 0.8604, a Recall of 0.9805, and a Precision of 0.7665 on the unseen test set. 
The comparative results are summarized in Table \ref{tab:main_results}. Our proposed model achieved the highest performance across Accuracy, Recall, and F1-score compared to traditional machine learning and deep learning architectures. It also outperformed the GCN-DP baseline across all evaluation metrics except Precision. This demonstrates that integrating geomorphological knowledge with deep learning frameworks enables superior feature capture for assessing topographic similarity.

\textbf{Validation of the Coarse Screening Strategy:}
\begin{figure}[htbp]
    \centering
    \begin{subfigure}[b]{0.45\textwidth}
        \centering
        \includegraphics[width=\textwidth]{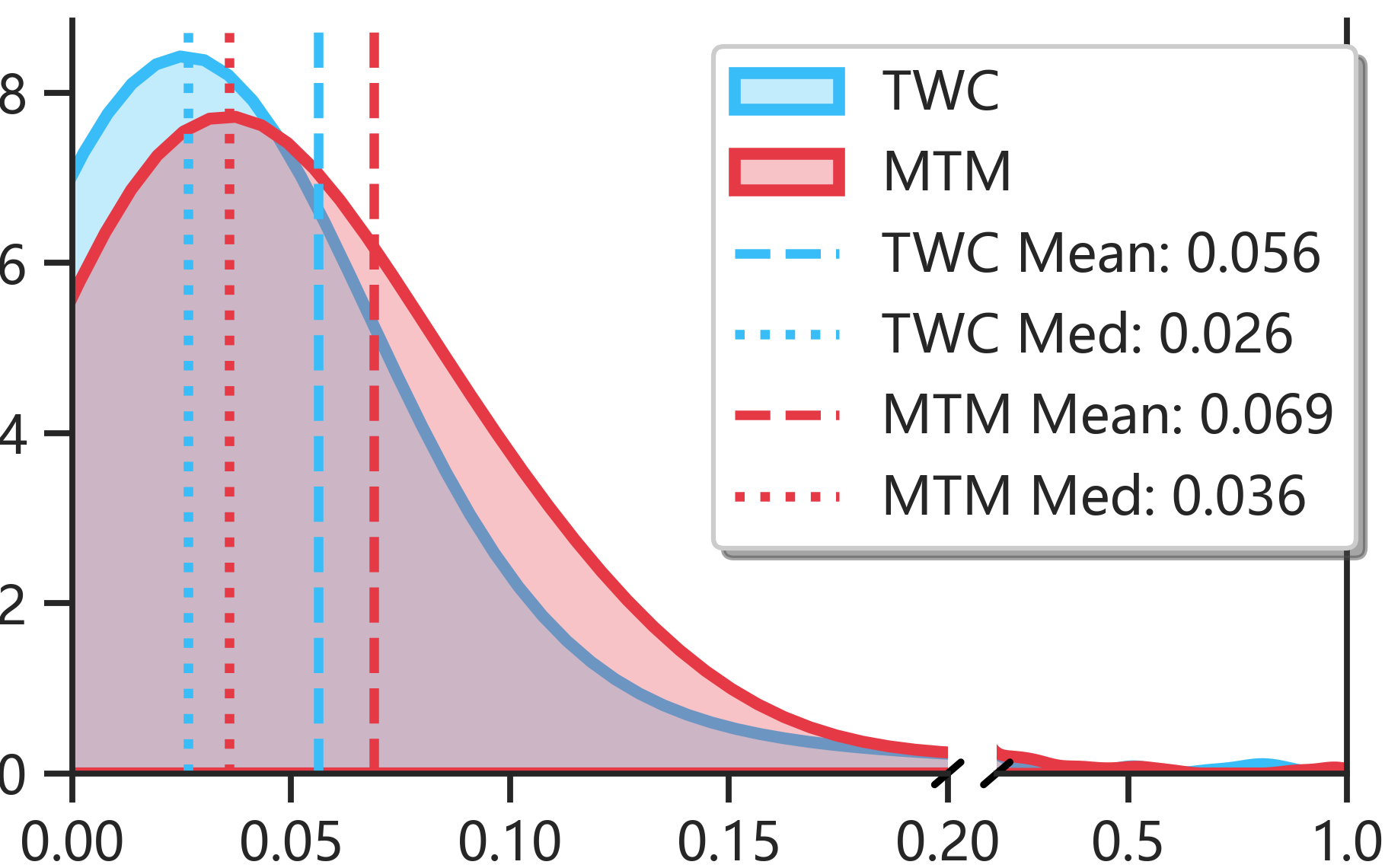} 
        \caption{Distributions of candidates eliminated by TWC and MTM respectively}
        \label{fig5sub1}
    \end{subfigure}
    \begin{subfigure}[b]{0.45\textwidth}
        \centering
        \includegraphics[width=\textwidth]{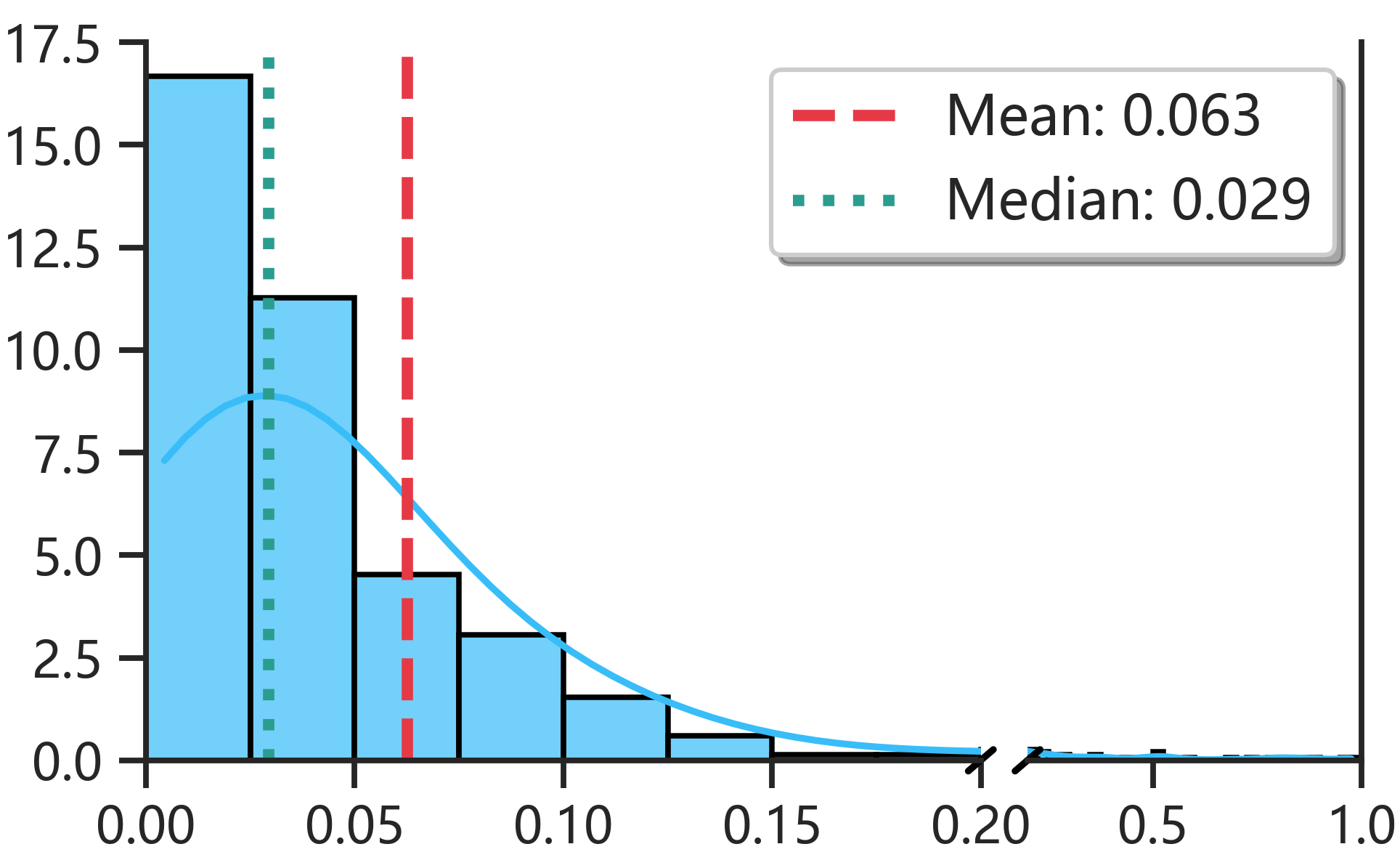}
        \caption{Combined distribution of all eliminated candidates}
        \label{fig5sub2}
    \end{subfigure}
    \caption{Topological similarity distributions of candidate valleys discarded during the coarse screening phase. (a) Separate distributions for the TWC and MTM stages. (b) The combined overall distribution.}
    \label{fig5}
\end{figure}
To validate the reliability of the TWC and MTM modules, the optimized model is deployed to compute the similarities between the reference trench and the candidate valleys previously filtered out by the TWC and MTM stages. As illustrated by the frequency distribution in Fig.~\ref{fig5sub1}, the eliminated valleys exhibit low similarity scores. Over 95\% of the valleys discarded by TWC had a similarity below 0.2, while for those discarded by MTM, over 95\% had a similarity below 0.21. For the overall valleys discarded in TWC and MTM (Fig.~\ref{fig5sub2}), the mean is 0.063, and the median is 0.029. These quantitative findings confirm that the filtered samples are topologically heterogeneous from the reference trench, thereby demonstrating the effectiveness and necessity of the TWC and MTM.

\textbf{Ecological Validity of Topographic Similarity:}
\begin{figure}[tbp]
    \centering
    \begin{subfigure}[b]{0.45\textwidth}
        \centering
        \includegraphics[width=\textwidth]{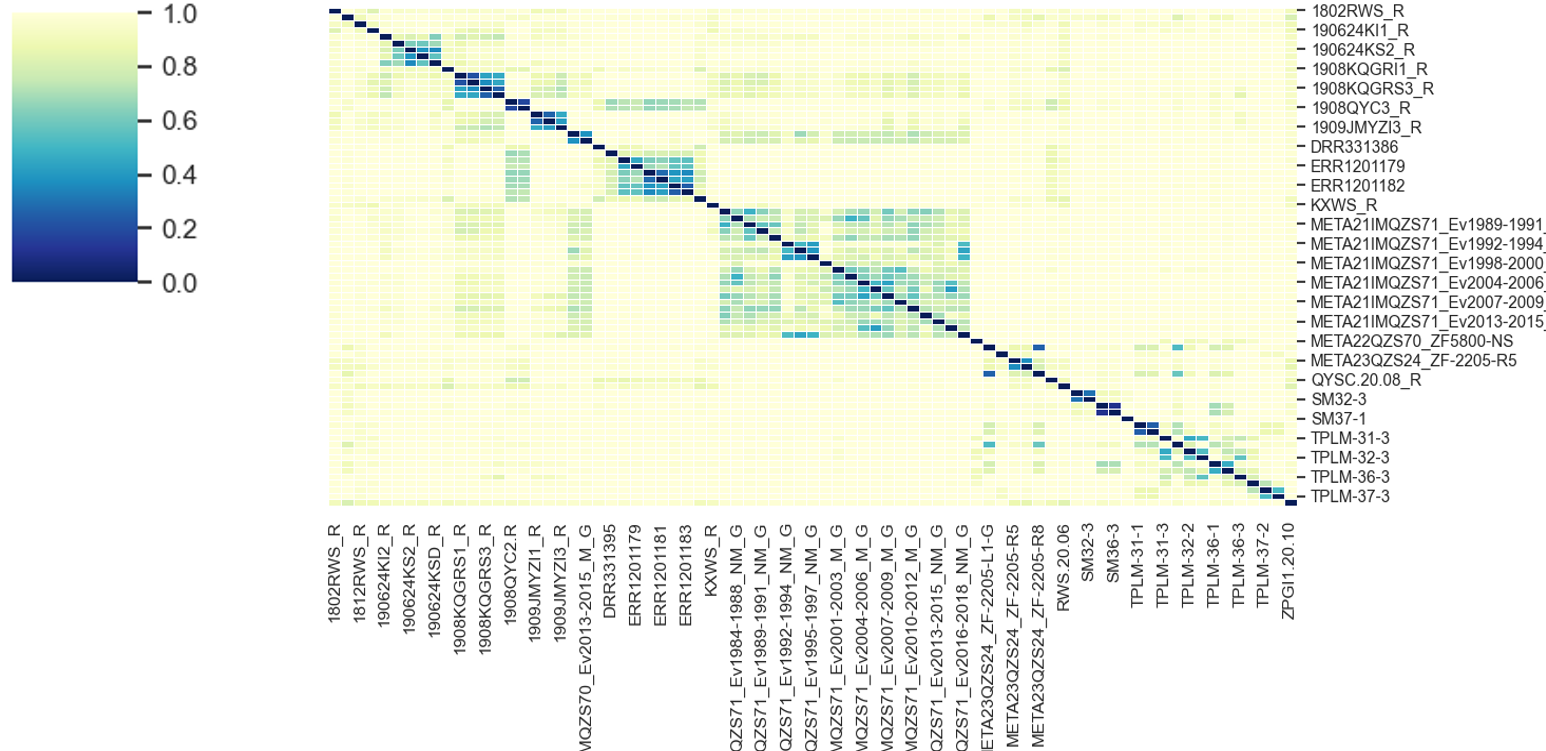}
        \caption{TP BC Dissimilarity Heat Map}
        \label{fig:top_left}
    \end{subfigure}
    \begin{subfigure}[b]{0.45\textwidth}
        \centering
        \includegraphics[width=\textwidth]{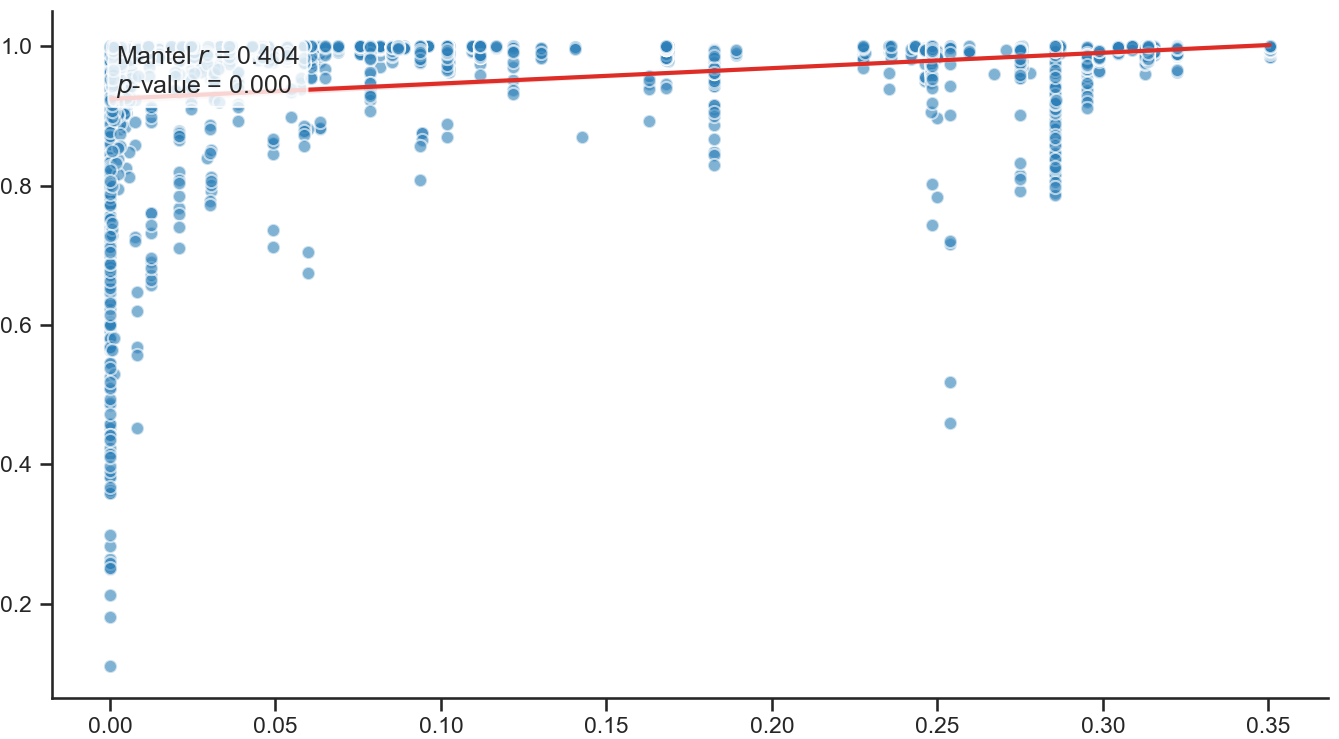} 
        \caption{TP Correlation between Terrain and Biological Dissimilarity}
        \label{fig:bottom_left}
    \end{subfigure}
    \caption{Bray-Curtis Dissimilarity heatmap and correlation between terrain and BC dissimilarity}
    \label{fig6}
\end{figure}
We also use the network to extract pairwise topological similarities among the candidate valleys, which are then converted into a topographic distance matrix. Concurrently, a Bray-Curtis dissimilarity matrix is constructed based on the relative abundances of biological communities at the corresponding sites (Fig.~\ref{fig:top_left}). Subsequently, a Mantel test is conducted to assess the correlation between topographic distance and biological dissimilarity (Figs.~\ref{fig:bottom_left}). The results indicate a highly significant positive correlation between the two matrices at the $p \le 0.01$ level. The strong correlation between terrain dissimilarity and BC dissimilarity indicates that our model has successfully captured topographic features; otherwise, such a strong correlation would be unlikely.

\subsection{Ablation Experiments}

\begin{table*}[h]
    \centering
    \caption{Ablation Study of MSG-Net}
    \label{tab:ablation_study}
    \renewcommand{\arraystretch}{1.2} 
    \begin{tabular*}{\textwidth}{@{\extracolsep{\fill}}lcccc}
        \toprule
        \textbf{Model Variant} & \textbf{Accuracy} & \textbf{Precision} & \textbf{Recall} & \textbf{F1-Score} \\
        \midrule
        \textbf{Full Features} & \textbf{0.8393} & \textbf{0.7665} & \textbf{0.9805} & \textbf{0.8604} \\
        \midrule
        w/o Con\_dense     & 0.7391 & 0.6571 & 1.0000 & 0.7931 \\
        w/o VRM            & 0.6522 & 0.5946 & 0.9565 & 0.7333 \\
        w/o ACR            & 0.6304 & 0.5750 & 1.0000 & 0.7302 \\
        w/o Slope          & 0.6739 & 0.6250 & 0.8696 & 0.7273 \\
        w/o Entropy        & 0.6522 & 0.5549 & 0.9565 & 0.7024 \\
        \bottomrule
    \end{tabular*}
\end{table*}
To evaluate the integrated geomorphological metrics, we ablated the five node attributes (Table \ref{tab:ablation_study}). The full MSG-Net achieves the highest F1-Score (0.8604). Removing any feature significantly degrades Precision (dropping up to 21.16\%) while keeping Recall disproportionately high ($\sim$1.0), indicating that missing physical priors makes the network over-sensitive to false positives.

Specifically, removing Contour Direction Shannon Entropy causes the steepest F1-Score drop (-0.1580), proving its vital role in encoding topographic complexity. Excluding Vector Ruggedness Measure or Arc-Chord Ratio reduces the F1-Score to $\sim$0.73, underscoring the necessity of fine-grained micro-surface extraction. Ablating macroscopic Slope severely degrades the F1-Score (0.7273) and triggers the largest Recall drop (to 0.8696), highlighting its function as an indispensable spatial baseline. Although Contour Density removal has the smallest impact—likely because the adjacency matrix inherently captures topological compactness—it remains a valuable supplement. Ultimately, these results confirm the five features act synergistically without redundancy.

\subsection{Visualization}
\begin{figure}[htbp]
    \centering
    \includegraphics[width=0.75\textwidth]{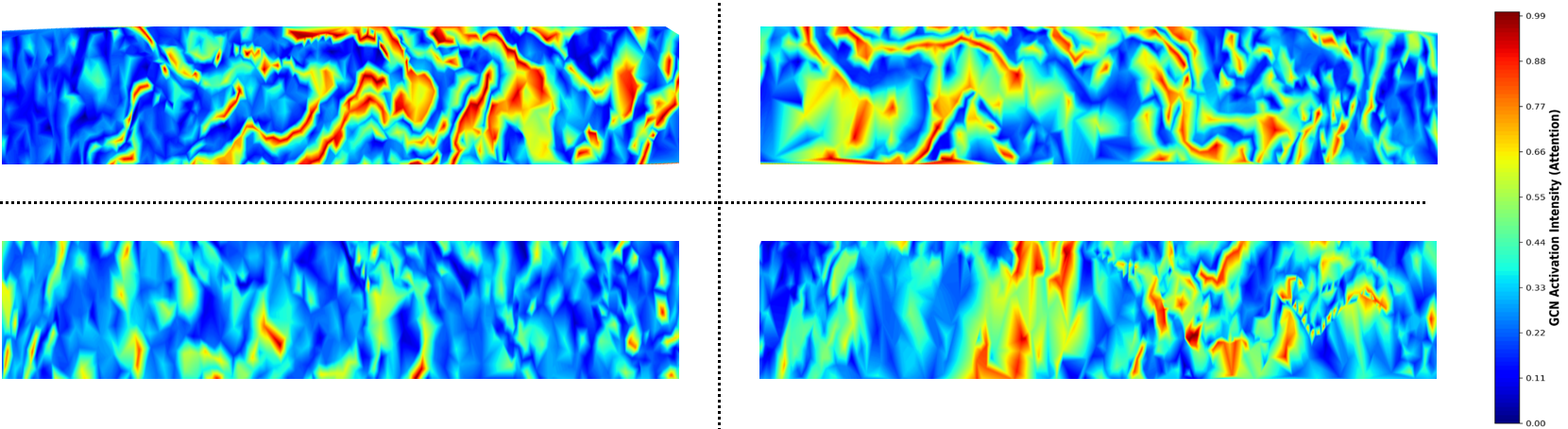}  
    \caption{Attention weight distribution map}
    \label{fig7}
\end{figure}
To demystify the internal decision-making mechanism of the MSG-Net and verify its geomorphological interpretability, we extract and visualize the neuronal activation intensities from the deepest layer of the GCN backbone. As illustrated in the attention weight distribution maps (Fig. \ref{fig7}), the highest attention weights (red hotspots) are consistently anchored in regions characterized by abrupt topographic transitions, such as ridges, valley bottoms, and complex terrain. These regions inherently possess high values of Vector Ruggedness Measure (VRM) and Arc-Chord Ratio (ACR). This demonstrates that the MSG-Net effectively learned to bypass redundant, uninformative planar spaces and focuses its computational resources on the distinct structural skeletons.

\section{Related Work}
Terrain representation is essential for similarity measurement and other geoscientific tasks.
Basic terrain representation includes two types: parametric representation converts terrain into geographically meaningful quantitative indicators (e.g., slope, aspect, elevation difference integral) \cite{pennock2003terrain,speight1977landform}; object-oriented representation characterizes terrain structure by extracting discrete geographic entities (e.g., watersheds, ridge lines, geomorphological units) \cite{jenson1988extracting,burrough2000principles}. Despite geographical significance and physical interpretability, these methods over-rely on isolated indicators or hand-crafted rules, failing to adequately capture overall terrain features. Thus, feature-based representation has become the mainstream for terrain similarity tasks, which converts terrain into abstract features that retain key morphological information\cite{MACLEOD200227}.

Feature-based methods capture abstract terrain morphology via diverse modeling approaches. Signal processing methods include: wavelet transform for multi-scale extraction of geomorphological and geological features with spatial localization \cite{kalbermatten2012multiscale,jordan2005application}; Fourier transform for extracting periodic global terrain features through frequency domain filtering \cite{gonzalez2025application}; Dynamic Time Warping (DTW) for flexible measurement of terrain waveform similarity and accommodation of terrain distortions \cite{mcclure2024integration}. For morphological decomposition, eigenshape analysis quantifies core morphological variations in terrain (e.g., symmetry, openness) by converting terrain contours into shape functions and extracting PCA-based feature vectors \cite{MACLEOD200227}.

With advances in computer vision and AI, data-driven terrain representation has emerged. Chen et al. \cite{chen2024terrain} proposed a CNN architecture fusing visual and proprioceptive signals for high-precision outdoor terrain classification; Li et al. \cite{li2020automated} realized automatic identification and localization of natural terrain features via improved CNN object detection; Ren et al. \cite{ren2024terrain} adopted Vision Transformer (ViT) to capture large-scale terrain structural correlations. Though they outperform traditional methods in accuracy and efficiency, these AI-based approaches lack support for geoscientific prior knowledge and oversimplify terrain into pixel grids.

To compensate for the lack of geographical thinking in pure AI models, academia has explored integrating geoscientific knowledge with AI. Feature-level fusion strategies supplement DEM data with terrain spatial attributes and environmental indicators \cite{zeng2022graph}; in-depth studies embed geographical priors into modeling mechanisms, such as Zhang et al. \cite{ZHANG2024323}, dividing slope units as graph nodes based on geographical environment similarity, and Kong et al. \cite{kong2025integrating}, constructing graph structures using terrain contour morphology.

\section{Conclusion}
In this paper, we present the \underline{\textbf{G}}eography-knowledge \underline{\textbf{E}}nhanced \underline{\textbf{A}}nalog \underline{\textbf{R}}ecognition (\textbf{GEAR}) Framework, a three-stage pipeline designed to recognize structurally homologous terrestrial analogs across extreme environments. By integrating geographical priors, GEAR effectively measures cross-domain topographic similarities, directly addressing the limitations of standard Computer Vision (CV) approaches. Empirical evaluations on the cross-validation dataset indicate that GEAR achieves an accuracy of 83.93\%, a recall of 98.05\%, and an F1-score of 86.04\%, outperforming the highest performing baseline, SANI-SSL (which achieved 79.82\% accuracy and 86.49\% recall).

Furthermore, this approach provides a computational framework for interdisciplinary geo-ecology to overcome the prohibitive costs of deep-sea biological sampling. GEAR allows researchers to leverage data from accessible environments to roughly estimate the biological conditions of the sub-seafloor and guide the selection of future sampling sites. Future work will integrate multi-modal variables (e.g., thermal gradients) into the graph representations and validate these biological predictions through subsequent oceanographic sampling.

\newcommand{\RoughSet}{\overline{\overline{V}}}  
\newcommand{\DTWSet}{\overline{V}}             
\newcommand{\FinalSet}{\dot{V}}                
\newcommand{\Target}{\mathcal{T}}              
\newcommand{\SimFunc}{\text{Sim}_{\text{TWC}}} 
\newcommand{\FeatVec}{\mathbf{Feat}_{\text{MTM}}} 
\newcommand{\ShapeFunc}{\mathcal{F}}           


\SetKwFunction{FDiff}{Diff}                  
\SetKwFunction{FDist}{EuclideanDist}         
\SetKwFunction{FResample}{Resample}          
\SetKwFunction{FFlat}{Flatten}               

\bibliographystyle{splncs04}
\bibliography{references}

\end{document}